\documentclass[runningheads,a4paper]{llncs}
\usepackage{makeidx}  
\usepackage{amssymb}
\usepackage{amsmath}
\usepackage{amsfonts}
\usepackage{graphicx}
\graphicspath{{figures/}}
\usepackage{multirow}
\usepackage{cite}
\usepackage{makecell}
\usepackage{float}
\usepackage[section]{placeins}
\begin{document}
\mainmatter

\title{Forming Predictive Features of Tweets for  Decision-Making Support}
\titlerunning{Forming Predictive Features of Tweets for  Decision-Making Support}

\author{Bohdan M. Pavlyshenko}

\authorrunning{Bohdan M. Pavlyshenko}

\institute{Ivan Franko National University of Lviv, Ukraine \\
b.pavlyshenko@gmail.com,  www.linkedin.com/in/bpavlyshenko/}

\maketitle              

\begin{abstract}
The article describes the approaches for forming different predictive features of tweet data sets and using them in the predictive analysis for decision-making support. 
The graph theory as well as frequent itemsets and association rules theory is used for  forming and retrieving different features from these datasests.  The use of  these approaches makes it possible to reveal a semantic structure in tweets related to a specified entity. 
 It is shown that quantitative characteristics of semantic frequent itemsets can be used in predictive regression models with specified target variables. 
\end{abstract}

\section{Introduction}

Tweets, the messages of Twitter microblogs,  have high density of semantically important keywords. It makes it possible to get semantically important information from the tweets and generate the features of predictive models for the decision-making support. Different studies of Twitter are considered in the papers~\cite{java2007we,kwak2010twitter,pak2010twitter, cha2010measuring,benevenuto2009characterizing,
bollen2011twitter,asur2010predicting,shamma2010tweetgeist, kraaijeveld2020predictive,
wang2020novel, balakrishnan2020improving}. 
In~\cite{pavlyshenko2019forecasting,pavlyshenko2019cantwitter}, we study the use of  tweet features for forecasting different kinds of events. 
In~\cite{pavlyshenko2020modelling}, we study the modeling of COVID-19 spread and its impact on the stock market using different types of data as well as  consider the features of tweets related to COVID-19 pandemic.

In this paper, we study the predictive features of tweets using loaded datasets of tweets related  to Tesla company. 

\section{Graph structure of tweets}
\begin{figure}
\center
\includegraphics[width=0.65\linewidth]{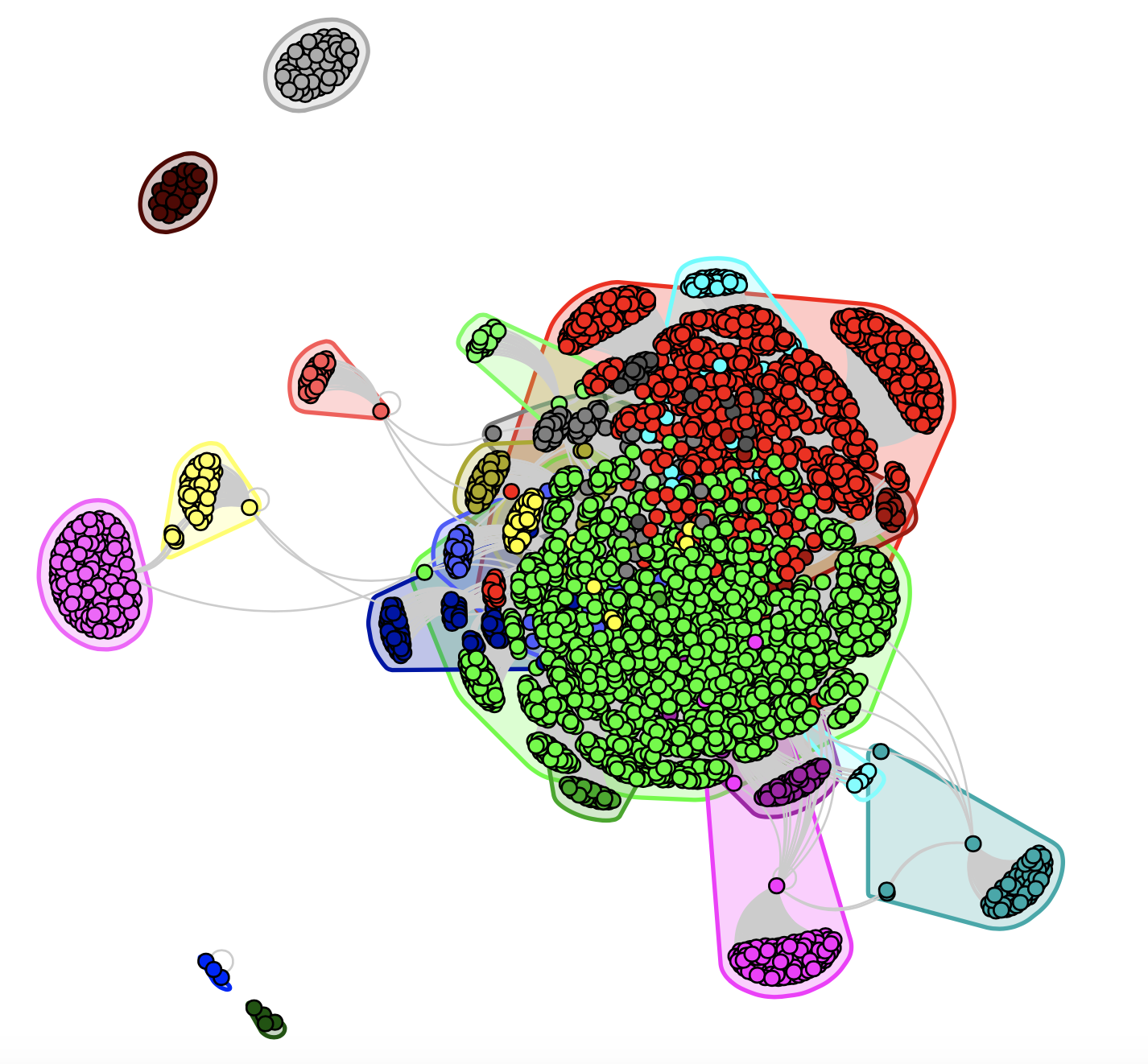}
\caption{Revealed users' communities for the subset of tweets}
\label{usr_graph}
\end{figure}

\begin{figure}
\centerline{\includegraphics[width=0.65\textwidth]{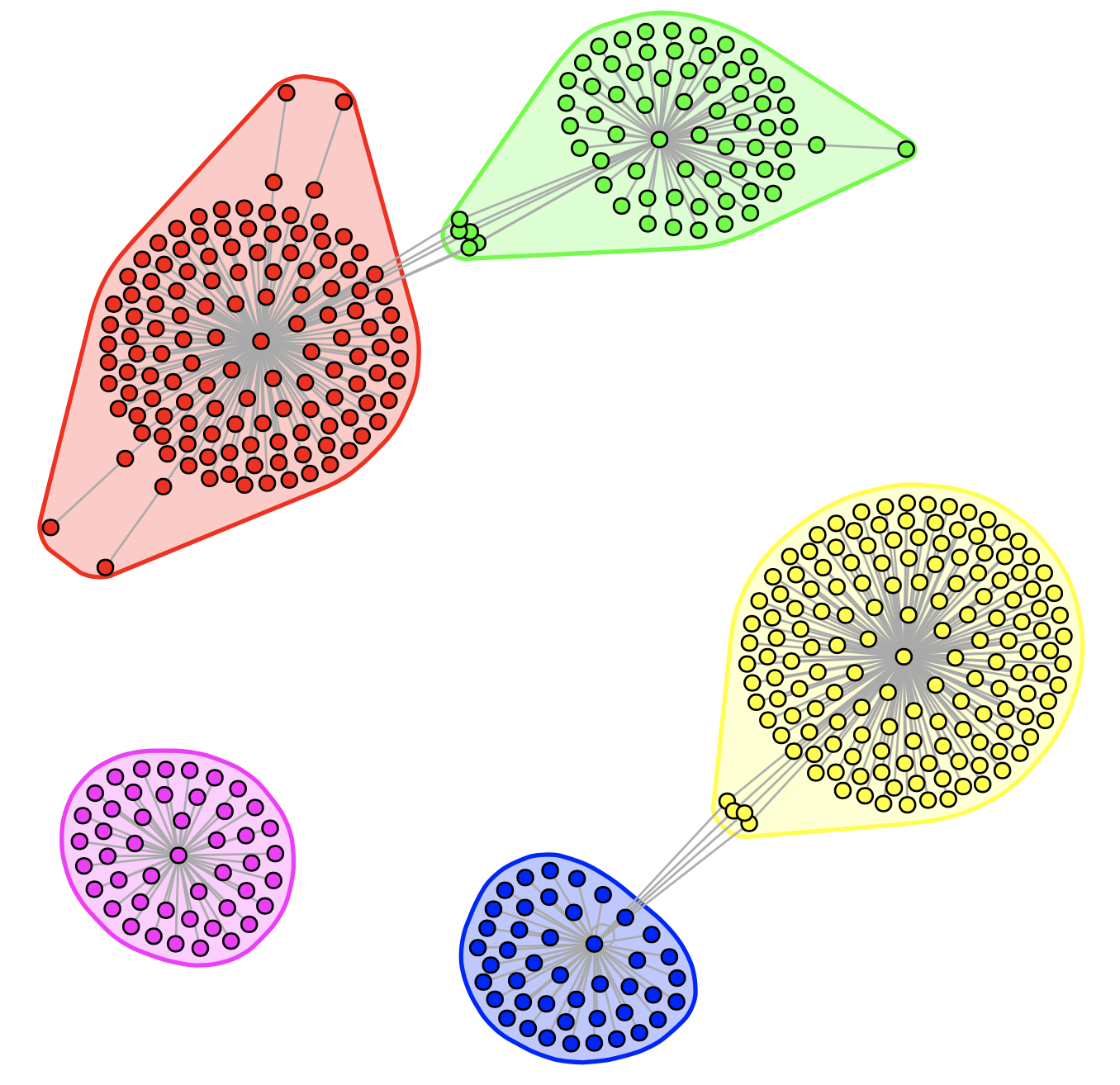}}
\caption{Subgraph for users of highly isolated communities}
\label{tesla_tw_p15}
\end{figure}

The  relationships among users can be considered as a graph, where vertices denote users and edges denote their connections.
Using graph mining algorithms, one can detect user communities and find ordered lists of users by various characteristics, such as
\textit {Hub, Authority, PageRank, Betweenness}. To identify user communities, we used the \textit{Community Walktrap Algorithm} algorithm, which is implemented in the package
 \textit{igraph}~\cite{csardi2006igraph} for the R programming language environment. We used the Fruchterman-Reingold algorithm from this package for visualization.
The \textit{Community Walktrap} algorithm searches for related subgraphs, also called communities, by random walk~\cite{pons2005computing}.
A graph which shows the relationships between users can be represented by
 Fruchterman-Reingold algorithm~\cite{fruchterman1991graph}.
We can assume that tweets could carry predictive information for different business processes. For our case study, we have loaded the tweets related to Tesla company for some time period. Qualitative structure can be used for aggregating different quantitative time series and, in such a way, creating new features for predictive models which can be used, for example, for stock prices forecasting. Let us consider which features we can retrieve from tweet sets for the predictive analytics. Figure~\ref{usr_graph} shows revealed users' communities for the subset of tweets.
Figure~\ref{tesla_tw_p15} shows the subgraph for users of highly isolated communities.
Revealing users' communities makes it possible to analyze different trends in tweet streams which are forming by different users' groups. 
\section{Analysis of tweets using frequent itemsets}
\begin{figure}
\center
\includegraphics[width=0.65\linewidth]{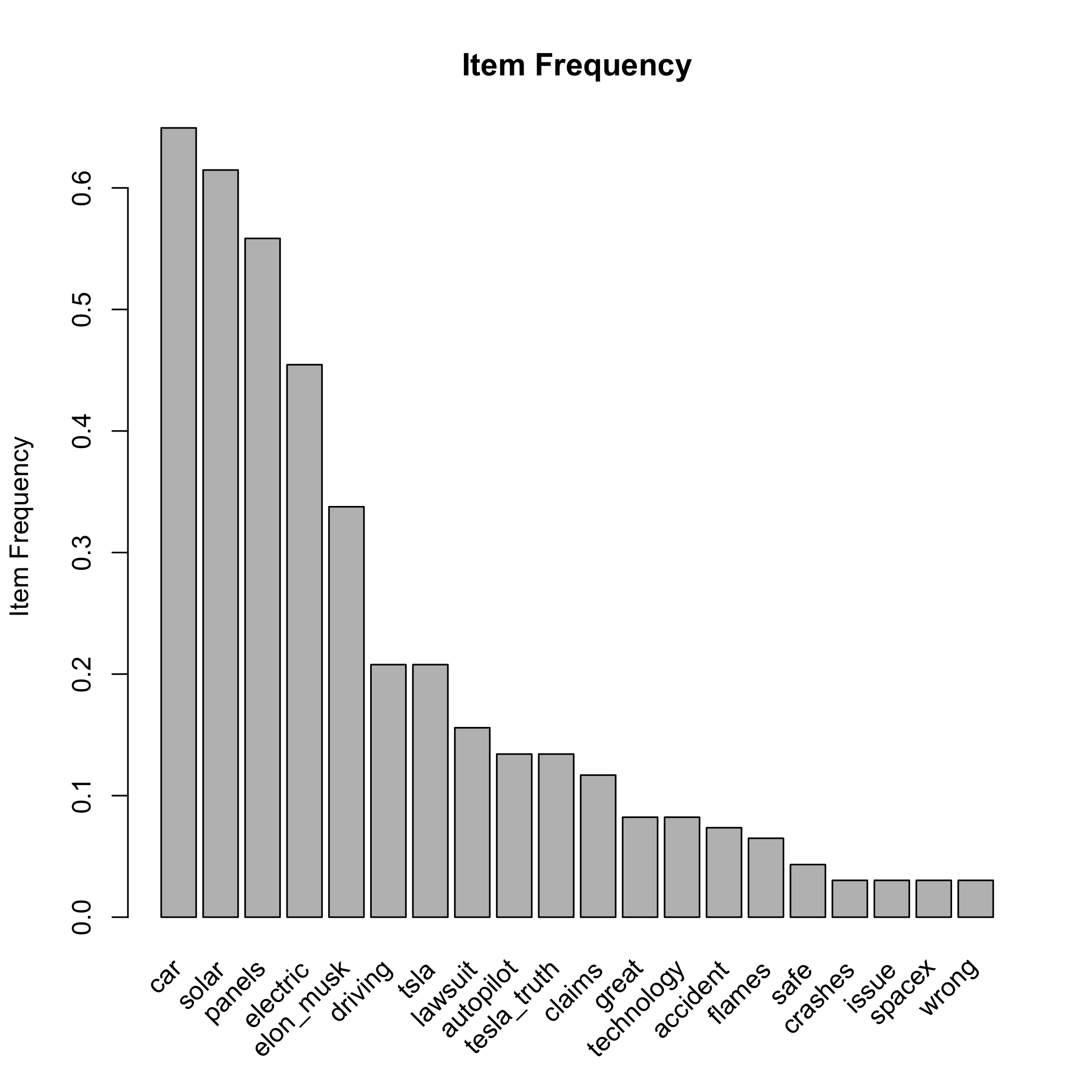}
\caption{Keyword frequencies for  the thematic field of frequent itemset analyis}
\label{tesla_freq}
\end{figure}

\begin{figure}
\centerline{\includegraphics[width=0.75\textwidth]{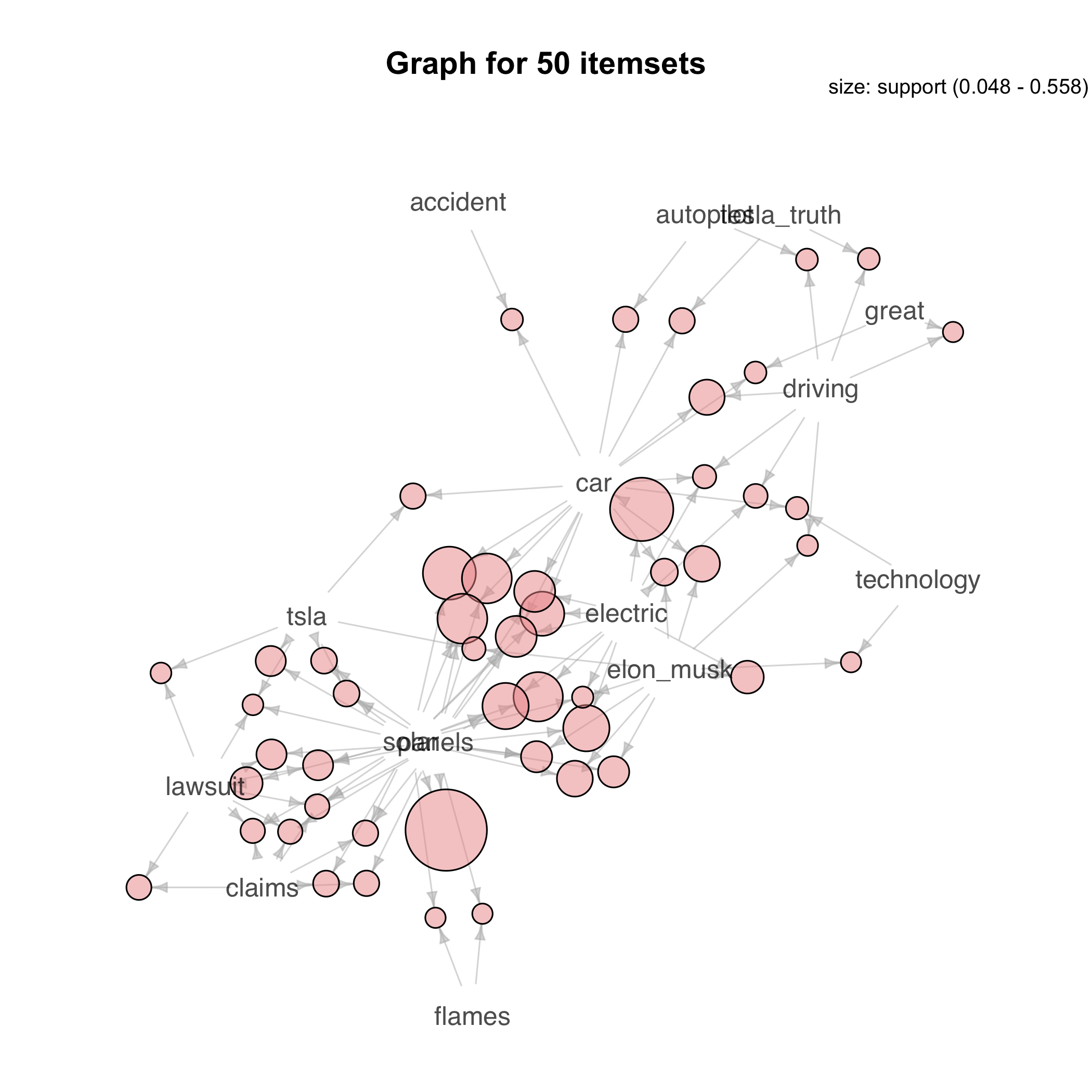}}
\caption{Semantic frequent itemsets}
\label{tesla_tw_p3}
\end{figure}

\begin{figure}
\centerline{\includegraphics[width=0.65\textwidth]{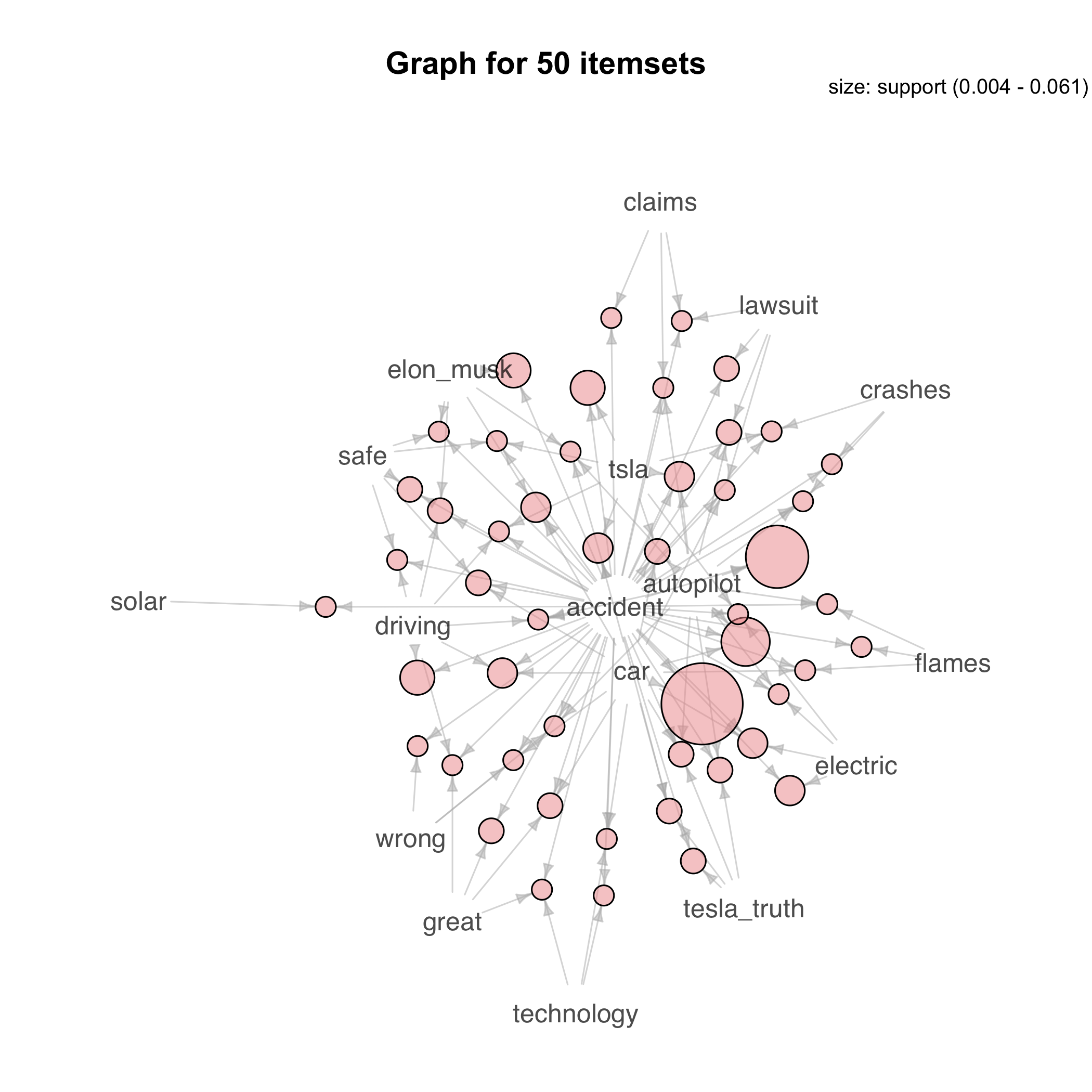}}
\caption{Semantic frequent itemsets}
\label{tesla_tw_p4}
\end{figure}

\begin{figure}
\centerline{\includegraphics[width=0.65\textwidth]{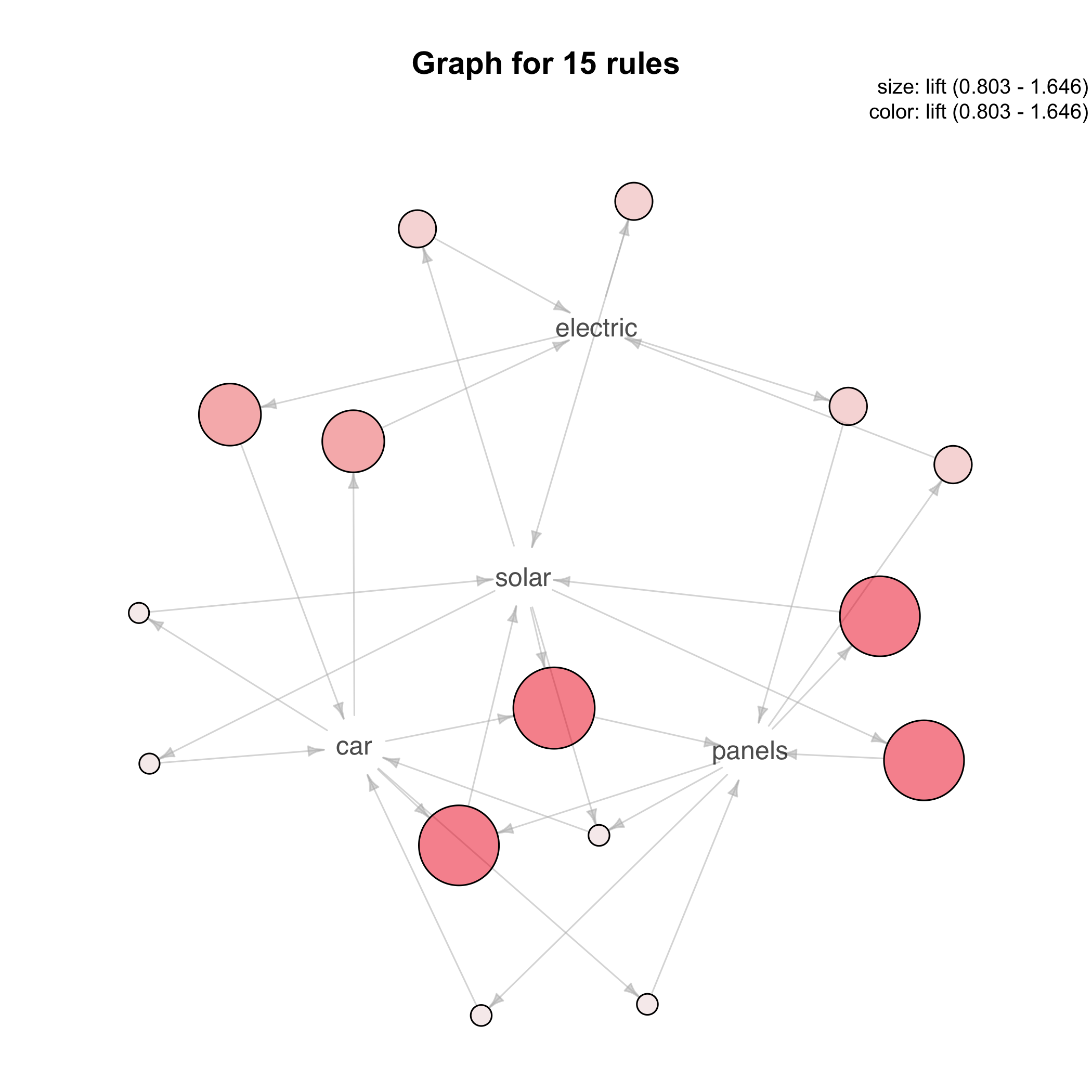}}
\caption{Associative rules, represented by a graph}
\label{tesla_tw_p5}
\end{figure}

\begin{figure}
\centerline{\includegraphics[width=0.65\textwidth]{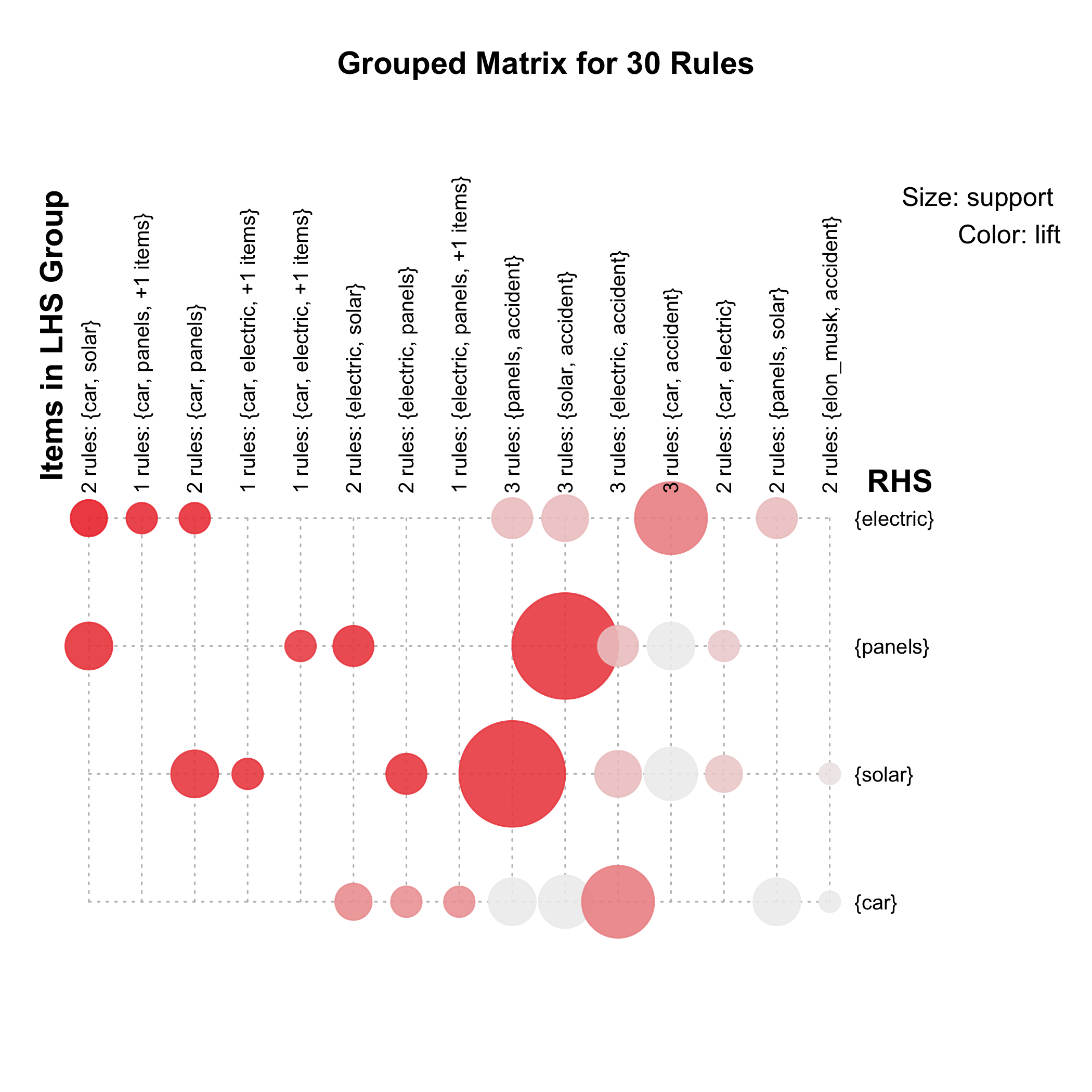}}
\caption{Associative rules represented by a grouped matrix}
\label{tesla_tw_p6}
\end{figure}

\begin{figure}
\center
\includegraphics[width=0.85\linewidth]{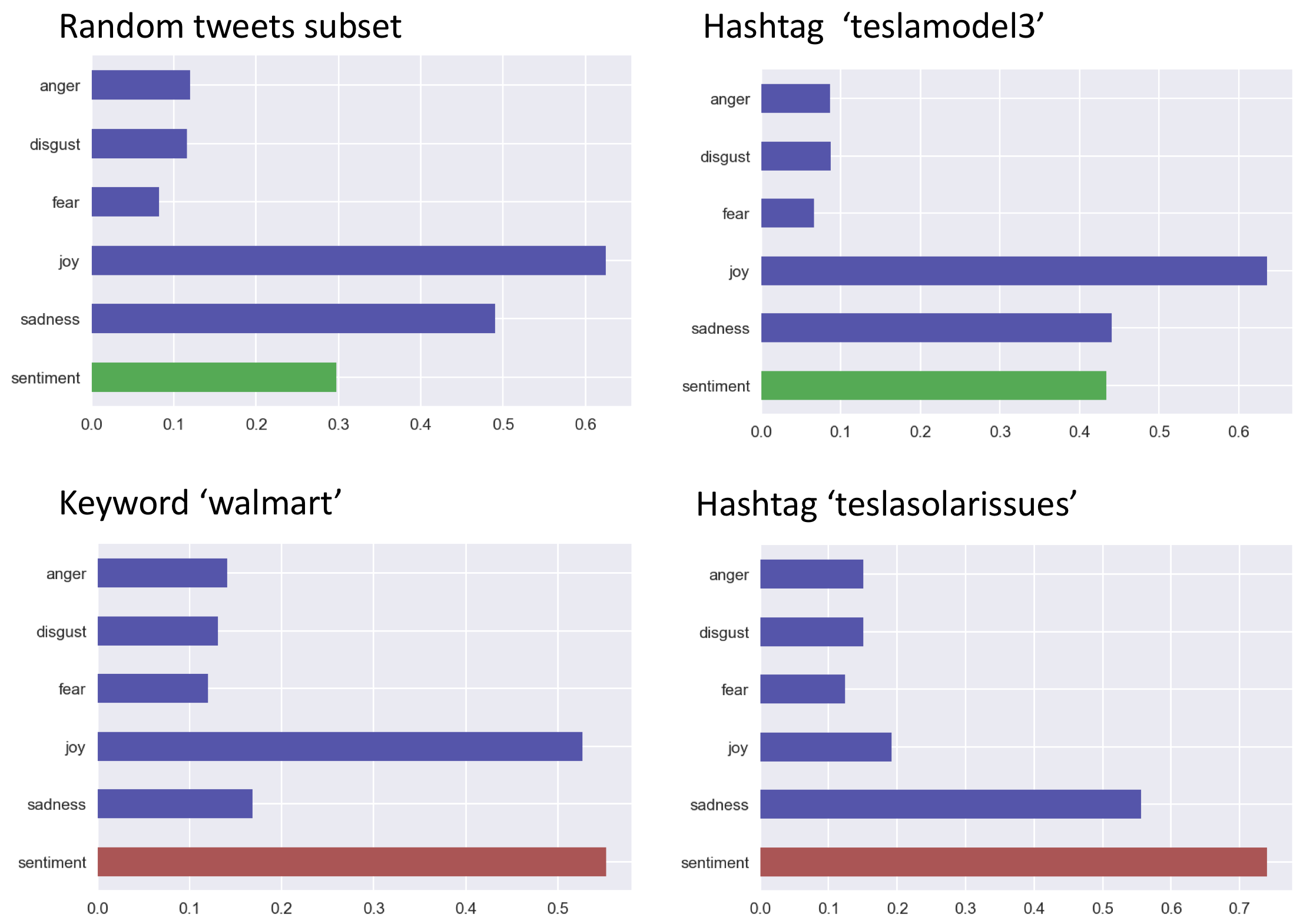}
\caption{Sentiment and personality analytics characteristics}
\label{sent_personalities}
\end{figure}

The frequent set  and associative rules theory is often used in the intelectual analysis~\cite{agrawal1994fast,agrawal1996fast,chui2007mining,gouda2001efficiently,
srikant1997mining,klemettinen1994finding,pasquier1999discovering,
brin1997beyond}.
 It can be used in a text data analysis to identify and analyze certain sets of objects, which are often found in large arrays and are characterized by certain features. 
Let's consider the algorithms for detecting frequent sets and associative rules on the example of processing microblog messages on Twitter. We can specify a thematic field which is a set of keywords semantically related to domain area under study.  Figure~\ref{tesla_freq} shows the frequencies of keywords   for  the thematic field of frequent itemsets analysis. This will make it possible to narrow the semantic analysis of messages to the given thematic framework. Based on the obtained frequent semantic sets, we are going to analyze  possible associative rules that reflect the internal semantic connections of thematic concepts in messages. In the time period when tweet dataset was being loaded, the accident with solar panels manufactured by Tesla on Walmart stores roofs took place. 
  It is important to consider the reflection of trends related to this topic in various processes, in particular, the dynamics of the company's stock prices in the financial market.
Using frequent itemsets and association rules, we can find a semantic structure in specified semantic fields of lexemes. 
Figures~\ref{tesla_tw_p3},~\ref{tesla_tw_p4}  shows semantic frequent itmesets for specified topics related to  Tesla company. Figures~\ref{tesla_tw_p5},~\ref{tesla_tw_p6} show association rules represented by graph and by grouped matrix.  
Figure~\ref{sent_personalities} shows sentiment and personality analytics characteristics received using IBM Watson Personality Insights~\cite{mahmud2016ibm}.
\section{Predictive analytics using tweet features}
\begin{figure}
\center
\includegraphics[width=1\linewidth]{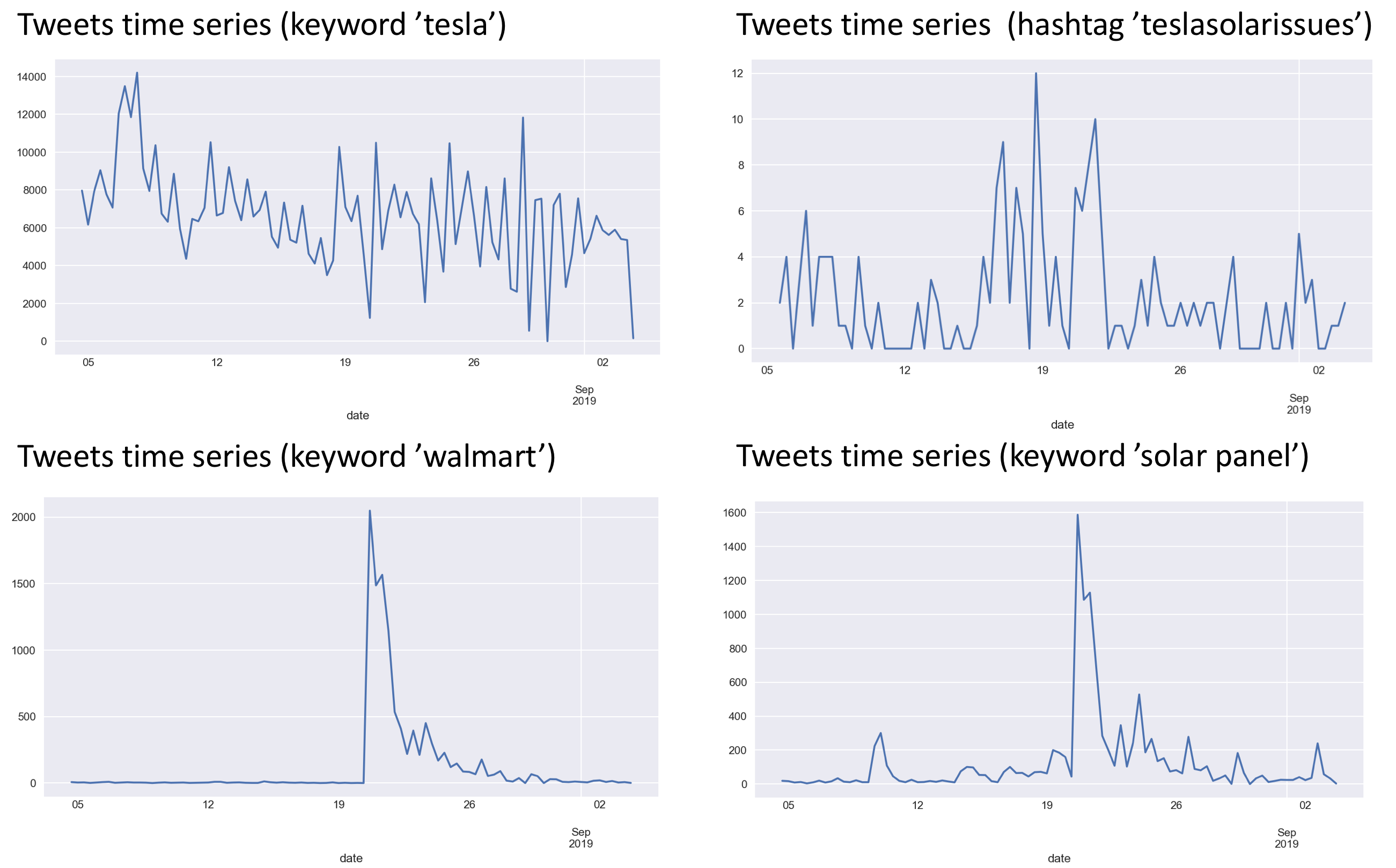}
\caption{Time series for different keywords and hashtags in the the tweets}
\label{kw_ts}
\end{figure}

\begin{figure}
\center
\includegraphics[width=1\linewidth]{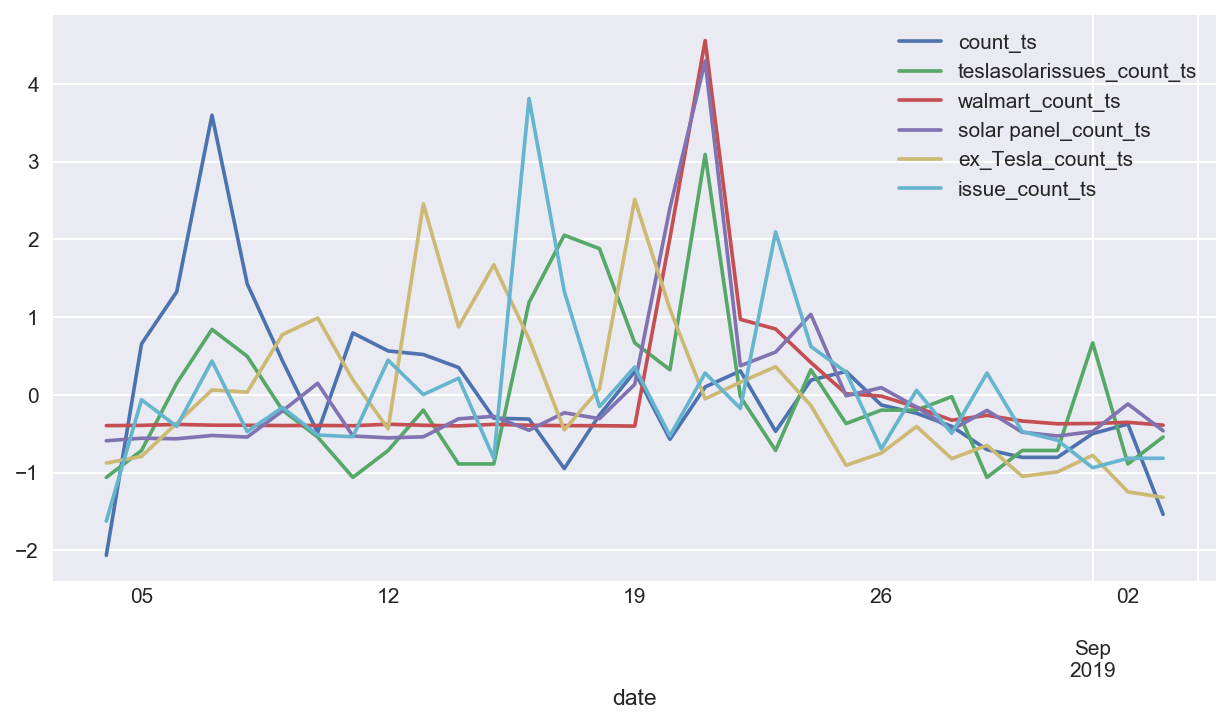}
\caption{Normalized keyword time series}
\label{tesla_kw_ts}
\end{figure}

\begin{figure}
\centerline{\includegraphics[width=0.85\textwidth]{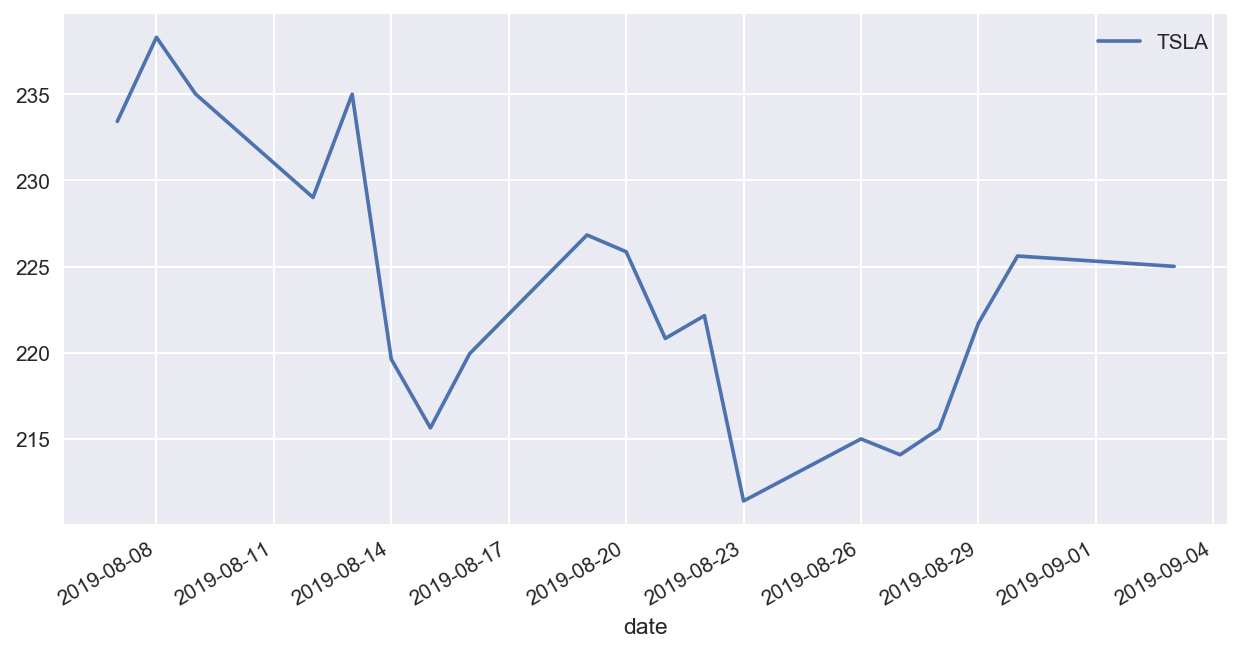}}
\caption{Dynamics of the stock price of \textit{Tesla} (TSLA ticker)}
\label{tesla_tw_p10}
\end{figure}

\begin{figure}
\center
\includegraphics[width=0.85\linewidth]{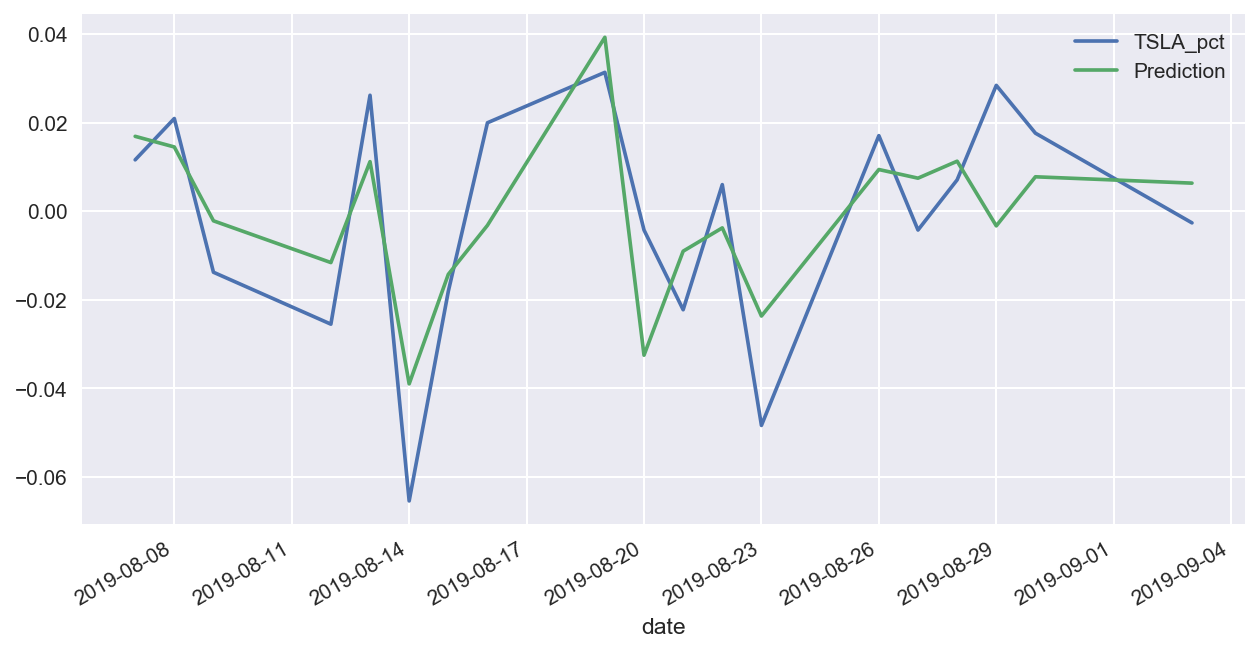}
\caption{Stock price return real and predicted values}
\label{tlsa_stock_ts_lasso_pred}
\end{figure}

\begin{figure}
\center
\includegraphics[width=0.75\linewidth]{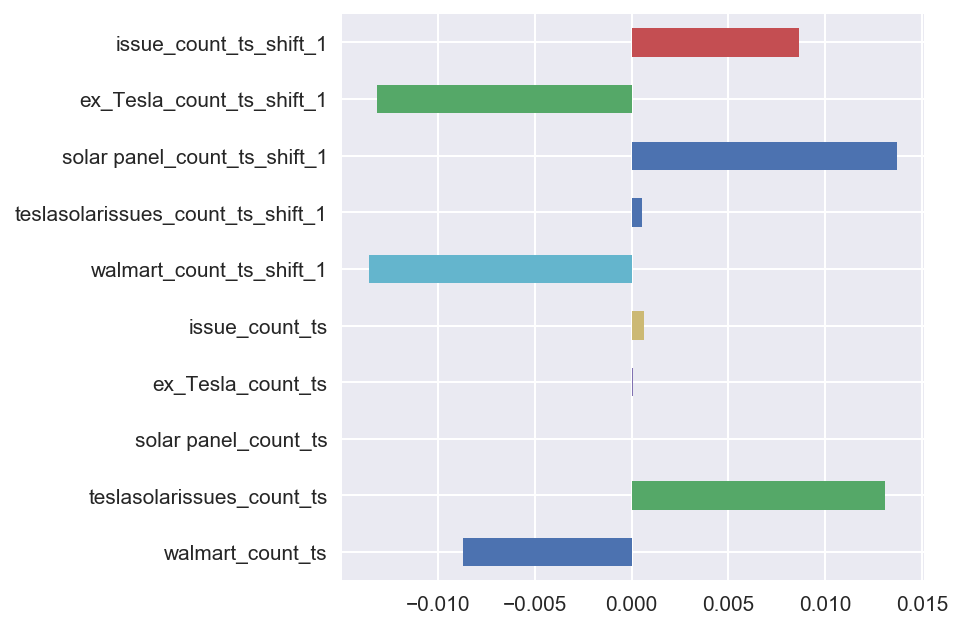}
\caption{Regression coefficients for the chosen features in the predictive model}
\label{tesla_stock_ts_lasso_coef}
\end{figure}

\begin{figure}
\center
\includegraphics[width=0.75\linewidth]{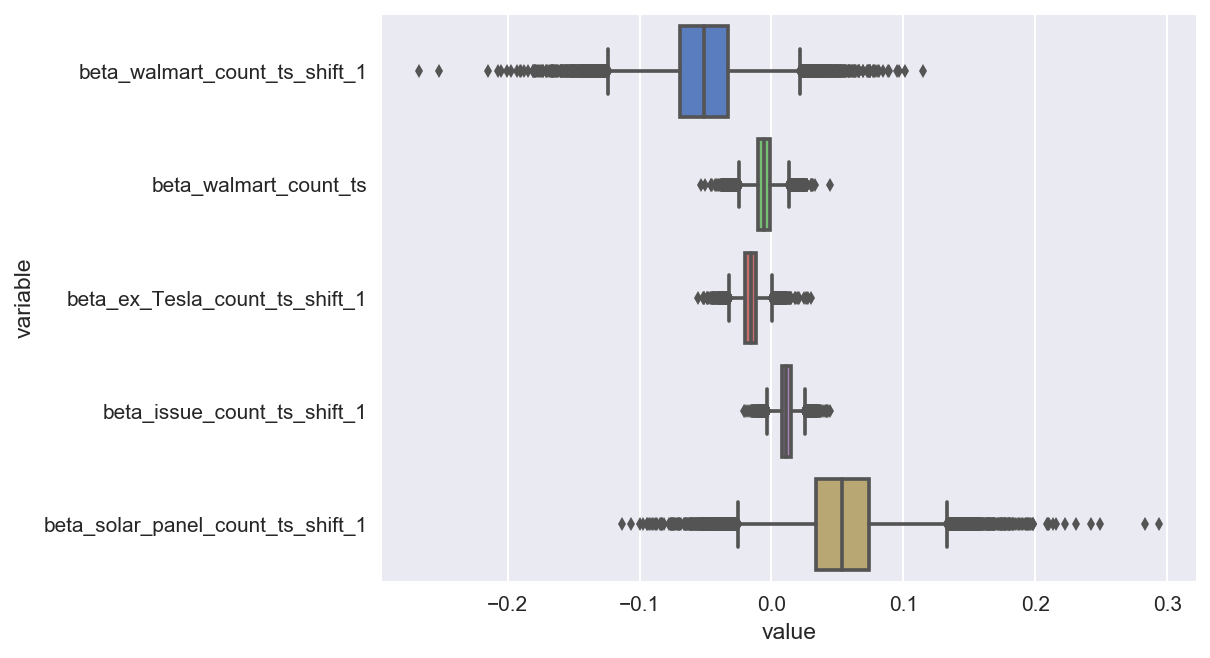}
\caption{Boxplots for feature coefficients in Bayesian regression model}
\label{boxplots_coef}
\end{figure}

Using revealed users' graph structure, semantic structure and topic related keywords and hashtags, one can receive keyword time series for tweet counts per day. These time series can be considered as features in the predictive models. In some time series, we can see when exactly the accident with solar panels on Walmart roof appeared and how long it was being considered in Twitter. Figure~\ref{kw_ts} shows the time series for different keywords and hashtags in the the tweets.
Figure~\ref{tesla_kw_ts} shows normalized keywords time series.
Social networks influence the formation of investment sentiment of potential stock market participants. Let us consider the dynamics of shares of the Tesla company  in the time period of the incident with solar panels  manufactured by Tesla.  It is reflected in the keywords time series on Figure~\ref{kw_ts}. One can see that at the time of the Tesla solar panel incident, the tweet activity is increasing over the time series of some keywords. Let us analyze how this incident affects the share price of Tesla. 
A linear model was created, where time series of keywords and their time-shifted values (lags) were considered as independent regression variables.
As a target variable, we considered the time series of the relative change in price during the day (price return).
Using LASSO regression, weights were found for the analyzed traits. 
Figure~\ref {tesla_tw_p10} shows the dynamics of the stock price Tesla (TSLA ticker) in the stock market.
We created a linear model where keyword time series and their lagged values were considered as covariates.  As a target variable, we considered stock price return time series for ticker TSLA. Using LASSO regression, we found weight coefficients for the features under consideration. Figure~\ref{tlsa_stock_ts_lasso_pred} shows the stock price return and predicted values.
Figure~\ref{tesla_stock_ts_lasso_coef} shows the regression coefficients for the chosen features in the predictive model.
We also conducted regression using Bayesian inference. Bayesian approach makes it possible to calculate the distributions for model parameters and for the target variable that is important for risk assessments~\cite{kruschke2014doing,gelman2013bayesian,carpenter2017stan}. 
Bayesian inference also makes it possible to  take into account non-Gaussian distribution of target variables that take place in many cases for financial time series. 
In~\cite{pavlyshenko2020bayesian}, we considered  different approaches of using Bayesian models for time series. Figure~\ref{boxplots_coef} shows the boxplots for feature coefficients in Bayesian regression model.
\section{Q-learning using tweet features}
\begin{figure}
\center
\includegraphics[width=0.85\linewidth]{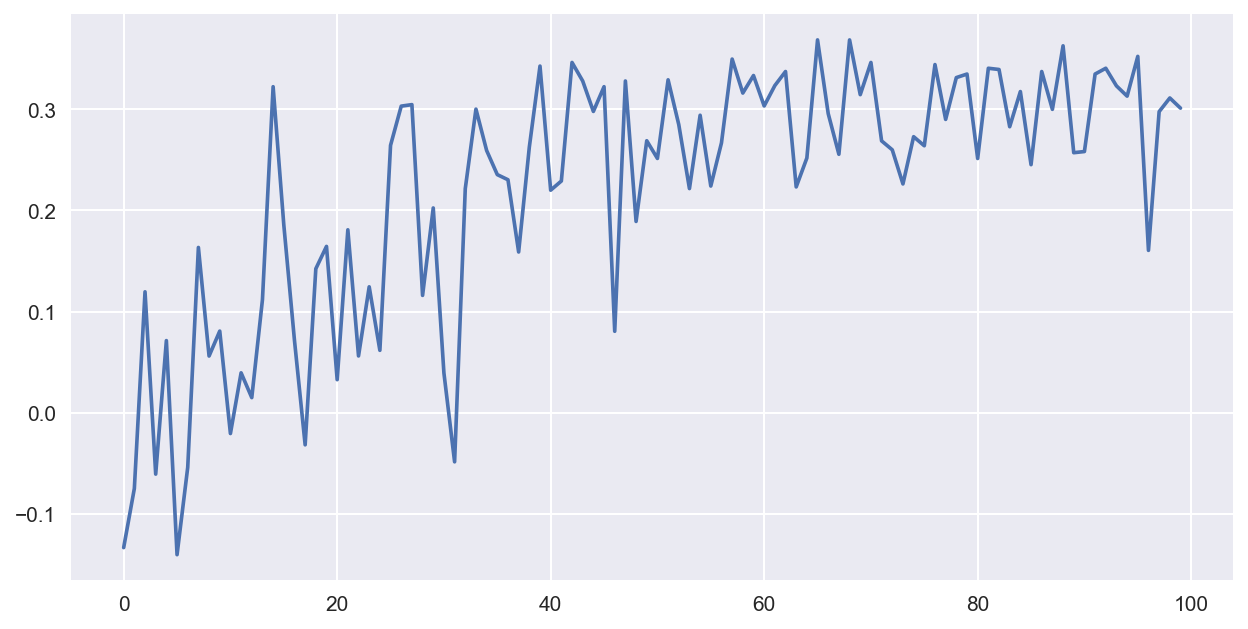}
\caption{Price return for the episodes for learning agent iterations}
\label{tsla_rl}
\end{figure}

It is interesting to use Q-learning to find an optimal trading strategy. 
Q-learning is an approach  based on the 
Bellman equation~\cite{sutton1998introduction,mnih2015human,mnih2013playing}.
In~\cite{pavlyshenko2020salests}, we considered different approaches for sales time series analytics using deep Q-learning.
Let us consider a simple trading strategy for the stocks with ticker TSLA.  In the simplest case of using deep Q-learning, we can apply three actions 'buy','sell','hold'. For state features, we used keyword time series. As a reward, we used stock price return. The environment for learning agent was modeled using keywords and reward time series. Figure~\ref{tsla_rl} shows the price return for the episodes for learning agent iterations.
The results show that an intelligent agent can find the an optimal profitable strategy. Of course, this is a very simplified case of analysis, where the effect of overfitting may occur, so this approach requires further study. The main goal is to show that, using reinforced learning and an environment model based on historical financial data and quantitative characteristics of tweets, it is possible to build a model in which an intelligent agent can find an optimal strategy that optimizes the reward function in episodes of  interaction of learning agent with the environment.  It was shown that time series of keywords features can be used as predictive features for different predictive analytics problems. Using Bayesian regression and tweets quantitative features one can estimate an uncertainty for the target variable that is important for the decision making support.
\section{Conclusion}
   Using the graph theory, the users' communities and influencers can be revealed given tweets characteristics. 
   The analysis of tweets, related to specified area, was carried out using frequent itemsets and association rules.
   Found frequent itemsets and association rules reveal the semantic structure of tweets related to a specified area. The quantitative characteristics of frequent itemsets and association rules, e.g. value of support, can be used as features in regression 
   models. Bayesian regression make it possible to assess the uncertainty of tweet features and target variable. It is shown 
   that tweet features  can also be used in deep Q-learning for forming the optimal strategy of learning agent e.g. in  the study of optimal trading strategies on the stock market.   
\FloatBarrier
\bibliographystyle{splncs04}
\bibliography{article.bib}
\end{document}